\documentclass{article}

\usepackage{microtype}
\usepackage{graphicx}
\usepackage{subfigure}
\usepackage{booktabs} % for professional tables

\usepackage{hyperref}

\usepackage[accepted]{icml2020}

\usepackage{amsmath}
\usepackage{amssymb}
\usepackage{amsfonts}
\usepackage{makecell}
\usepackage{multirow}
\usepackage{outlines}

\icmltitlerunning{An Empirical Study of Invariant Risk Minimization}

\begin{document}

\twocolumn[
\icmltitle{An Empirical Study of Invariant Risk Minimization}



\icmlsetsymbol{equal}{*}

\begin{icmlauthorlist}
\icmlauthor{Yo Joong Choe}{kb}
\icmlauthor{Jiyeon Ham}{kb}
\icmlauthor{Kyubyong Park}{kb}
\end{icmlauthorlist}

\icmlaffiliation{kb}{Kakao Brain, Seongnam-si, Gyeonggi-do, South Korea}

\icmlcorrespondingauthor{Yo Joong Choe}{yj.choe@kakaobrain.com}

\icmlkeywords{Invariant Risk Minimization, Out-of-Distribution Generalization}

\vskip 0.3in
]



\printAffiliationsAndNotice{}  % leave blank if no need to mention equal contribution

\begin{abstract}
Invariant risk minimization (IRM) \cite{arjovsky2019invariant} is a recently proposed framework designed for learning predictors that are invariant to spurious correlations across different training environments.
%Because IRM does not assume that the test data is identically distributed as the training data, it can allow models to learn invariances that generalize well on unseen and out-of-distribution (OOD) samples.
Yet, despite its theoretical justifications, IRM has not been extensively tested across various settings. 
In an attempt to gain a better understanding of the framework, we empirically investigate several research questions using IRMv1, which is the first practical algorithm proposed to approximately solve IRM.
By extending the ColoredMNIST experiment in different ways, we find that IRMv1 (i) performs better as the spurious correlation varies more widely between training environments, (ii) learns an approximately invariant predictor when the underlying relationship is approximately invariant, and (iii) can be extended to an analogous setting for text classification.
%We hope that this work will shed light on the characteristics of IRM and help with applying IRM to real-world OOD generalization tasks.
\end{abstract}

\section{Introduction}

Invariant risk minimization (IRM) \cite{arjovsky2019invariant} is a recently proposed machine learning framework where the goal is to learn invariances across multiple training environments \cite{peters2016causal}. 
Compared to the widely used framework of empirical risk minimization (ERM), IRM does \emph{not} assume that training samples are identically distributed.
Rather, IRM assumes that training samples come from multiple environments and seeks to find associations that are invariant across those environments.
This allows its resulting predictor to be effective in out-of-distribution (OOD) generalization, i.e., achieving low error on test samples that might come from an unseen environment.

Although IRM is a promising framework for OOD generalization, it is not extensively tested across many settings in which IRM is expected to perform well. 
Experiments in \cite{arjovsky2019invariant} are limited to two-environment binary classification tasks, which do not cover different types of multi-environment settings found in real-world datasets \cite{torralba2011unbiased,beery2018recognition,geva2019modeling}.
%We believe that the lack of empirical validations makes it difficult to apply IRM to real-world tasks that require OOD generalization. 

In this paper, we conduct a series of experiments that examine the extent to which the IRM framework can be effective.
Specifically, we extend the ColoredMNIST setup from \cite{arjovsky2019invariant} in several ways and compare the OOD generalization performances of ERM and IRMv1, which is the first practical algorithm for IRM proposed by \cite{arjovsky2019invariant}.
We find that:
\vspace{-1em}
\begin{itemize}
\setlength{\itemsep}{0pt}
\setlength{\parskip}{0pt}
\setlength{\parsep}{0pt} 
    %\item The generalization performance of IRMv1 improves as the difference between training environments, in terms of the degree of spurious correlations, becomes larger. (\S \ref{sec:q1})
    \item The generalization performance of IRMv1 improves as the training environments become more diverse. (\S \ref{sec:q1}) %as the spurious correlation varies more widely between training environments. (\S \ref{sec:q1})
    %\item IRMv1 works even when the invariant correlation is stronger than the spurious correlation. (\S \ref{sec:q2})
    \item IRMv1 learns an approximately invariant predictor when the underlying relationship is approximately invariant. (\S \ref{sec:q3})
    %\item IRMv1 can be applied to multiple environments (\S \ref{sec:q4}) and output classes (\S \ref{sec:q5}). 
    \item IRMv1 can be extended to an analogous setup for a text classification task. (\S \ref{sec:nlp})
\end{itemize}
\vspace{-0.5em}

We publicly release our code at {\url{https://github.com/kakaobrain/irm-empirical-study}}.

\section{Background: Invariant Risk Minimization}
\label{sec:irm}

\subsection{The IRM Framework}
\label{sec:framework}

Consider a set of environments $\mathcal{E}$. 
For each environment $e \in \mathcal{E}$, we assume a data distribution $\mathcal{D}^e$ on $\mathcal{X} \times \mathcal{Y}$ and a risk function $R^e(f) = \mathbb{E}_{(X^e, Y^e) \sim \mathcal{D}^e} [L(f(X^e), Y^e)]$ for a convex and differentiable loss function $L$, such as cross-entropy and mean squared error. 
Our goal is to find a predictor $f: \mathcal{X} \to \mathcal{Y}$ that minimizes the maximum risk over all environments, or the \textbf{OOD risk}: $R^{\text{OOD}}(f) = \max_{e \in \mathcal{E}} R^e(f)$.

A representation function $\Phi: \mathcal{X} \to \mathcal{H}$ is said to elicit an \textbf{invariant predictor} $w \circ \Phi$ if there exists a classifier $w: \mathcal{H} \to \mathcal{Y}$ that is simultaneously optimal for all environments, i.e. $w \in \arg\min_{\bar{w}} R^e(\bar{w} \circ \Phi)$ for all $e \in \mathcal{E}$. 
Given training environments $\mathcal{E}_\text{tr} \subseteq \mathcal{E}$, \textbf{IRM} learns an invariant predictor $w \circ \Phi$ by solving the following bi-level optimization problem:
\begin{align}\label{eqn:irm}
    \min_{\Phi, w}\quad &\sum_{e \in \mathcal{E}_\text{tr}} R^e(w \circ \Phi) \\
    \text{subject to}\quad & w \in \arg\min_{\bar{w}} R^e ( \bar{w} \circ \Phi)\;\; \forall e \in \mathcal{E}_\text{tr} \nonumber
\end{align}

\citet{arjovsky2019invariant} use the theory of invariant causal prediction (ICP) \cite{peters2016causal} to illustrate that, for linear predictors and a linear underlying causal structure, an invariant predictor can be found using IRM as long as the training environments are sufficiently diverse and the underlying invariance is satisfied. 
The connection to ICP also yields a causal interpretation of IRM: it can discover causal structures from the underlying data distribution that can be extrapolated to OOD datasets.

\subsection{The IRMv1 Algorithm}
\label{sec:irmv1}

Because \eqref{eqn:irm} is highly intractable, particularly when $\Phi$ is allowed to be non-linear, \citet{arjovsky2019invariant} proposed a tractable variant that approximates \eqref{eqn:irm}, called \textbf{IRMv1}:
\begin{align}\label{eqn:irmv1}
    \min_{\Phi} &\sum_{e \in \mathcal{E}_\text{tr}} \left[ R^e(w \cdot \Phi) + \lambda \cdot || \nabla_{w | w = \mathbf{1} } R^e (w \cdot \Phi) ||_2^2 \right]
\end{align}
where $\Phi: \mathcal{X} \to \mathbb{R}^{k}$ for output dimension $k$, $w \in \mathbb{R}^{k}$ is an $k$-dimensional vector, and $\mathbf{1}$ is an $k$-dimensional all-1 vector.
$\lambda \in [0, \infty)$ is a hyperparameter that balances between predictive power over training tasks, i.e., the ERM loss, and the squared gradient norm, i.e., the IRM penalty.
Note that, in \cite{arjovsky2019invariant}, the IRMv1 derivation focused on the case where $k = 1$, where the predictor outputs a single scalar (e.g., a logit).
Throughout this paper, we build predictors that output $k \geq 2$ dimensional logits for $k$-class classification, such that the gradient in the penalty term is also $k$-dimensional. 

Importantly, the penalty term captures how much the invariant representation $\Phi$ can be improved by locally scaling itself. 
For each $\Phi$ and $e$, the squared gradient norm penalty approximates ``how close'' $w$ is to a minimizer $w_\Phi^e \in \arg\min_{\bar{w}} R^e(\bar{w} \circ \Phi)$.
%Indeed, for strongly convex loss functions and full-rank representations $[\Phi(X_i^e)]_{i=1}^n$, the penalty is zero if and only if $w$ is the unique minimizer $w_\Phi^e$. 
%If the penalty term is zero across training environments, then the ``classifier'' $w$ is simultaneously optimal for all training environments, making $w \circ \Phi$ an invariant predictor among training environments. 
Note that \eqref{eqn:irmv1} is now an objective with respect to $\Phi$ only and can be optimized using gradient descent for non-linear $\Phi$, such as a neural network.
%Section 3 of \cite{arjovsky2019invariant} details how \eqref{eqn:irmv1} approximates \eqref{eqn:irm} in the case of linear least-squares regression given representations $\Phi(X^e)$ and targets $Y^e$.

\section{Extended ColoredMNIST}
\label{sec:coloredmnist}

In this section, we examine IRMv1 on different versions of Extended ColoredMNIST, a collection of synthetic image classification tasks derived from MNIST.
Extended ColoredMNIST tweaks the original ColoredMNIST setup from \cite{arjovsky2019invariant}, where the training set of MNIST is split into two environments and spurious correlations are introduced by associating specific colors with specific output classes.
The overall pipeline for constructing a Extended ColoredMNIST dataset can be summarized as follows.

\textbf{Extended ColoredMNIST: Data Construction Pipeline} 
\vspace{-1em}
\begin{enumerate}
\setlength{\itemsep}{0pt}
\setlength{\parskip}{0pt}
\setlength{\parsep}{0pt}
    \item Randomly split the training data ($n=50,000$) into $m$ environments, $e_1, \dotsc, e_m$. The test data ($n=10,000$) is considered to come from its own environment $e_\mathrm{test}$.
    \item Corrupt each label with probability $\eta_e$.
    \item Pair each output class with a unique color, e.g., $(\mathrm{class}_1, \mathrm{color}_1)$, $(\mathrm{class}_2, \mathrm{color}_2)$, and so on.
    \item With probability $p_e$, color the input image with the color paired with its (possibly corrupt) label. Otherwise, color the input image with a different color.
\end{enumerate}
\vspace{-0.5em}

The goal of this setup, which includes\footnote{See Appendix \ref{app:original} for a detailed explanation.} the original ColoredMNIST construction \cite{arjovsky2019invariant}, is to build a set of training environments where the {\bf spurious correlation} is stronger than the {\bf invariant correlation} in data, but only the spurious correlation varies among the training environments.
The spurious correlation is captured by the correlation between color and label, or $1 - p_e$, while the invariant correlation is captured by the correlation between shape and label, or $1 - \eta_e$.
In this setting, we expect that ERM picks up the spurious correlation that appears strong during training and suffers in the test set where the spurious correlation is altered. 
On the other hand, we expect that IRM picks up the invariant correlation only, because it learns an invariant predictor. 
%Note that the original ColoredMNIST setup from \cite{arjovsky2019invariant} is a special case of Extended ColoredMNIST.\footnote{See Appendix \ref{app:original} for a detailed explanation.}

In the following experiments, we use the training configurations used in the original implementation\footnote{\url{https://github.com/facebookresearch/InvariantRiskMinimization}}, unless noted otherwise. 
Full training details are presented in Appendix \ref{app:coloredmnist_hparams}.
Due to space constraints, we highlight two of our findings for Extended ColoredMNIST here; we include additional experimental results in Appendix \ref{app:coloredmnist_additional}.

\subsection{How does the diversity of training environments affect IRMv1's performance?}
%\subsection{How does the difference between training environments affect IRMv1's performance?}
\label{sec:q1}

First, we examine how the OOD generalization performance of IRMv1 is affected by the difference between training environments in terms of their spurious correlations.
For $m=2$ training environments, the gap between training environments is captured by $|p_1 - p_2|$, the difference in coloring probabilities.
Since IRM exploits this gap in spurious correlations among the environments, we expect that the algorithm can perform well only if this gap is substantially greater than zero.
Note that, if $p_1 = p_2$, then no algorithm can distinguish the invariant correlation from the (no longer) spurious one.

\begin{figure}[t]
\begin{center}
\centerline{\includegraphics[width=1\columnwidth]{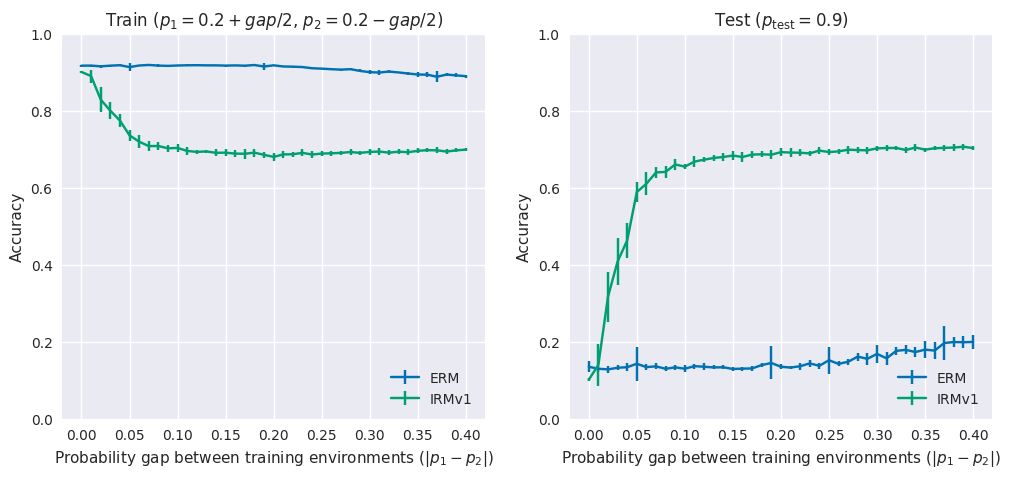}}
\caption{Accuracy on Extended ColoredMNIST, train (left) and test (right), \emph{versus the difference in spurious correlations between the two training environments, $|p_1 - p_2|$.} Averaged over 10 trials (error bars represent standard deviations).}
\label{fig:q1}
\end{center}
\end{figure}

In Figure \ref{fig:q1}, we plot the train and test accuracies of ERM and IRMv1 against $|p_1 - p_2|$. 
For each value of $gap := |p_1 - p_2|$, we define $p_1 = 0.2 + gap/2$ and $p_2 = 0.2 - gap/2$. 
Since the average of coloring probabilities is $\frac{p_1+p_2}{2} = 0.2$, the spurious correlation is always stronger than the invariant correlation during training.
All other settings are the same as ColoredMNIST -- in particular, a random baseline achieves 50\% accuracy and the optimal classifier achieves 75\%.

As shown in the right plot of Figure \ref{fig:q1}, we find that \emph{the generalization performance of IRMv1 consistently improves as the gap between training environments grows larger.} 
Note that IRMv1 outperforms ERM as soon as a non-zero gap exists and achieves above-chance accuracy for $|p_1 - p_2| \geq 0.05$.
As the gap grows larger, IRMv1's test set accuracy consistently improves and eventually gets close to the accuracy (70.6\%) of the grayscale model, which is the same model trained on data without spurious correlations (i.e., the colors). 
This suggests that IRMv1 benefits from more varying spurious correlations between training environments.

\subsection{How does approximate invariance affect IRMv1's performance?}
\label{sec:q3}

In ColoredMNIST and in \S\ref{sec:q1}, shape and label were associated with the same degree of correlation ($1-\eta_e$) for all environments. 
%In the original ColoredMNIST and in \S\ref{sec:q1}, the association between shape and label was \emph{completely} invariant --- shape and label were correlated with the same degree for each environment.
% --- 75\% with label corruption and 100\% without across all environments.
However, in reality, it may be unreasonable to assume that this association is \emph{completely} invariant. 
\citet{arjovsky2019invariant} states that, when the data follows an approximately invariant model, IRM should return an approximately invariant solution, because its objective is a differentiable function with respect to the training environments. 
Our goal here is to empirically validate this claim.

We set up a different version of Extended ColoredMNIST, where we now vary the rate of label corruption across environments. 
Specifically, we evaluate IRMv1 and ERM on different values of $|\eta_1 - \eta_2|$, while keeping their average fixed to the test set, i.e., $\frac{\eta_1+\eta_2}{2} = \eta_\mathrm{test} = 0.25$. 
We also fix $p_1 = 0.2$ and $p_2 = 0.1$.
As $|\eta_1 - \eta_2|$ grows larger, the association between shape and label becomes less invariant.

\begin{figure}[t]
\begin{center}
\centerline{\includegraphics[width=1\columnwidth]{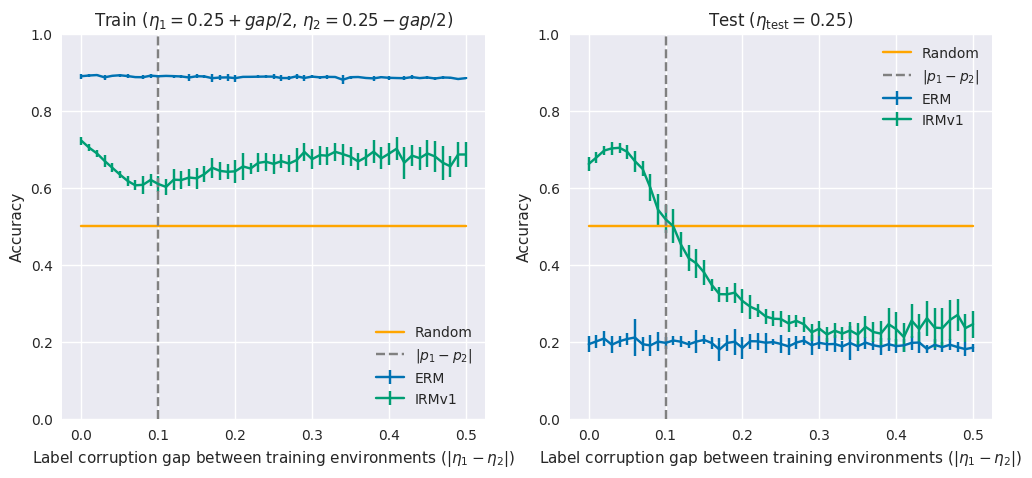}}
\caption{Accuracy on Extended ColoredMNIST, train (left) and test (right), \emph{versus the gap in label corruption ratio across training environments, $|\eta_1 - \eta_2|$.} Averaged over 10 trials (error bars represent standard deviations).}
\label{fig:q3}
\end{center}
\end{figure}

Figure \ref{fig:q3} shows mean accuracies of ERM (blue), IRMv1 (green), and random (orange) predictors across different values of $|\eta_1 - \eta_2|$ (0.0 to 0.5 with 0.01 increments). 
First, we find that \emph{IRMv1 can achieve a high test accuracy even when the association between shape and label is only approximately invariant.}
For example, when $|\eta_1 - \eta_2| = 0.05$, IRMv1 achieves a 70.4\% test accuracy, which is even larger than the test accuracy when $\eta_1 = \eta_2$ (66.2\%).
As long as the association is not more variant than the spurious correlation between color and label, IRMv1 performs above chance and also well above ERM, which quickly overfits to color across all values of $|\eta_1 - \eta_2|$.

Interestingly, the overall behavior of test accuracy for IRMv1 (Figure \ref{fig:q3}, right) suggests that \emph{the algorithm gives more weight to whichever factor that is more invariant, in a nearly smooth manner.}
As the gap grows larger, IRMv1's test accuracy becomes directly correlated to $|\eta_1 - \eta_2|$: high when the gap is small and low when the gap is large. 
In fact, this pattern in IRMv1 test accuracy is almost a smooth function of the gap, starting from high test accuracy (i.e., predictions made off of shape) eventually reaching the test accuracy of ERM (i.e., predictions made off of color).
Also notable is that the accuracy is close to the random baseline (50\%) precisely when $|\eta_1 - \eta_2| = |p_1 - p_2| = 0.1$.
We posit that, at this point, the algorithm either discards both factors, as they are equally non-invariant, or weighs both factors equally, which would cancel out in terms of test accuracy.
In either case, it is interesting to see that IRMv1 chooses not to favor color, which is a stronger indicator of the label than shape in the training set.

\section{PunctuatedSST-2}
\label{sec:nlp}

As described in \cite{arjovsky2019invariant}, the fact that ERM is prone to absorbing biases and spurious correlations from the training data is a fundamental problem across all machine learning applications.
Natural language processing (NLP) is no exception: several recent papers \cite{gururangan2018annotation,mccoy2019right,niven2019probing} repeatedly pointed out the presence of spurious correlations in text classification tasks, often in the form of specific words being highly correlated with specific labels, and how NLP models actively exploit them.
As a result, state-of-the-art models for NLP often make trivial mistakes and fail to generalize out-of-distribution.

With this in mind, we apply the data construction pipeline for Extended ColoredMNIST to a text classification dataset and evaluate the performances of both ERM and IRMv1.
We start with Stanford Sentiment Treebank (SST-2) \cite{socher2013recursive}, a standard benchmark dataset for binary sentiment analysis, and use an analogous pipeline.

\textbf{PunctuatedSST-2: Data Construction Pipeline}
\vspace{-1em}
\begin{enumerate}
\setlength{\itemsep}{0pt}
\setlength{\parskip}{0pt}
\setlength{\parsep}{0pt}    
    \item Randomly split the SST-2 training data ($n=67,350$) into $m$ environments, $e_1, \dotsc, e_m$. The SST-2 validation data ($n=873$) is \emph{also} randomly split, but into $m+1$ environments: $e_1, \dotsc, e_m$ and $e_\text{OOD}$.
    \item Corrupt each label with probability $\eta_e$.
    \item Pair each output class with a punctuation mark: positive with a period (.) and negative with an exclamation mark (!).
    \item Remove any existing punctuation mark at the end of each input sentence.
    \item With probability $p_e$, punctuate the input sentence with the mark paired with its (possibly corrupt) label. Otherwise, punctuate the input sentence with the other mark.
\end{enumerate}
\vspace{-0.5em}

There are two key differences from the pipeline for Extended ColoredMNIST.
First, we now keep a separate test set that is assumed to have come from each of the training environments.
This allows us to more precisely measure how much the model is picking up the spurious correlation, as the examples from this test set are unseen but has the same degree of spurious correlation as its corresponding training environment.
%In particular, a model's test set accuracy from environment $e_i$ cannot surpass $1 - \eta_i$ \emph{unless} the model pays attention to the spurious correlation.
The final test set, now renamed as coming from environment $e_\text{OOD}$, corresponds to the test set we had in Extended ColoredMNIST.\footnote{The same could have been done for Extended ColoredMNIST, but we preserved the original pipeline by \cite{arjovsky2019invariant} to make our results comparable.}
Second, because the inputs are now sequences of tokens, we need a different way to ``color'' them based on the label. 
In PunctuatedSST-2, we introduce associations between each output class and a specific punctuation mark. 
Adding a punctuation mark at the end allows us to preserve both the meaning and grammaticality of the input sentence.
Aside from these two differences, we use the original ColoredMNIST setup for the following experiments: $m=2$ training environments with $p_1 = 0.2$, $p_2 = 0.1$, and $p_\mathrm{OOD} = 0.9$, and label corruption across environments of $\eta_1 = \eta_2 = \eta_\mathrm{OOD} = 0.25$.

\subsection{Can IRMv1 work with spurious token-to-label correlations in text classification tasks?}
%\subsection{Results with a Bag-of-Words Model}

Because state-of-the-art models on SST-2 can be quite complex for our tests, we resort to a simple bag-of-words (BOW) model with averaged word embeddings \cite{pennington2014glove}.
We still train using gradient descent, although we run a new hyperparameter search \cite{bergstra2015hyperopt}.
Further training details can be found in Appendix \ref{app:punctuatedsst2_hparams}.

\begingroup
\setlength{\tabcolsep}{4pt} % Default value: 6pt
\begin{table}[t]
\centering
%\large
\begin{tabular}{c | r r r}
\Xhline{1.1pt}
\multirow{2}{*}{\bf Algorithm} & \multicolumn{3}{c}{\bf Test Accuracy} \\
                           & \multicolumn{1}{c}{$e_1$} & \multicolumn{1}{c}{$e_2$} & \multicolumn{1}{c}{$e_\mathrm{OOD}$}         \\ \hline
ERM                        & $71.2 \pm 0.6$ & $81.8 \pm 3.2$ & $30.4 \pm 1.6$ \\ 
IRMv1                      & $57.7 \pm 1.8$ & $62.1 \pm 2.2$ & $\bf 61.4 \pm 2.2$ \\ \hline
Majority                   & $50.3$         & $50.2$         & $52.2$         \\
Oracle                     & $58.7 \pm 2.0$ & $65.8 \pm 2.4$ & $61.5 \pm 1.2$ \\ 
\Xhline{1.1pt}
\end{tabular}
\caption{Test accuracy (\%) on PunctuatedSST-2. Averaged over 10 trials (mean $\pm$ standard deviation). $e_1$ and $e_2$ refer to held-out data sampled from the same distributions as the two training environments ($p_1=0.2$, $p_2=0.1$). $e_\mathrm{OOD}$ refers to a held-out set with an inverted spurious correlation ($p_\mathrm{OOD}=0.9$).
}
\label{tbl:punctuatedsst2}
\end{table}
\endgroup

In Table \ref{tbl:punctuatedsst2}, we report the mean accuracies for ERM and IRMv1, as well as for the majority vote and oracle (ERM without input perturbations, or the ``grayscale'' model) baselines.
We first find that the ERM model is highly susceptible to the spurious token-to-label correlations present in the training environments, resulting in a worse-than-chance OOD accuracy (30.4\%) as in ColoredMNIST.
More importantly, we find that \emph{the IRMv1 penalty works to remove the effect of this spurious but varying correlation, achieving almost as high OOD accuracy (61.4\%) as the oracle model (61.5\%)}\footnote{The bag-of-words model is known to only achieve $\sim$80\% accuracy before label corruption \cite{wang2018glue}. Because we also add a 25\% label corruption, we can expect even the oracle model to have an even lower accuracy.}.
% Note that, compared to models for ColoredMNIST, the BOW model is known to only achieve around 80\% accuracy before label corruption \cite{wang2018glue}. 
% As a result, after label corruption ($\eta=0.25$), the oracle model's accuracy for PunctuatedSST-2 (around 61\%) are lower than that of ColoredMNIST (around 71\%). 
These results suggest that the IRMv1 algorithm can indeed be extended to analogous settings in NLP.

\section{Conclusion and Future Work}
\label{sec:conclusion}

To deepen our understanding of IRM, we examined the generalization performance of IRMv1 across several extensions of ColoredMNIST.
Overall, our findings were optimistic. 
We found that IRMv1 is capable of detecting and removing small variations across environments and that it learns approximately invariant predictors when the underlying relationship is approximately invariant.
We also found that IRMv1 can be extended to analogous text classification tasks involving spurious token-to-label correlations.

We believe that these results serve as initial steps toward making IRM more broadly and effectively applicable. 
One important future direction is to identify multi-environment settings where we can apply IRM to achieve OOD generalization, or more specifically, how we find or construct training environments that are sufficiently diverse.
\citet{teney2020unshuffling}, for example, recently illustrated a use case in visual question answering (VQA), where question types or (learned) clusters in inputs were used to create different training environments.
% One important future direction is to figure out what kinds of multi-environment settings exist in real-world datasets and how the IRM framework can help.
% This includes identifying patterns of varying spurious correlations across different tasks, such as the word-label correlations we described in text classification.
% \citet{teney2020unshuffling} recently showed interesting applications of the IRM framework on visual question answering, but identifying applicable scenarios across and beyond vision and NLP tasks remains an open problem.
% Another, perhaps more challenging, direction is to identify ways to build meaningful multi-environment settings.
% Even though nature does not shuffle data \cite{arjovsky2019invariant}, many standard benchmark datasets already come in a single-environment manner, making it infeasible to distinguish invariant and spurious correlations.
% It would be important to have better insights on both how to effectively construct multi-environment datasets and how to identify multiple environments within an existing dataset.
Developing more stable approximations of the IRM framework, as pointed out in \cite{ahuja2020invariant} and \cite{teney2020unshuffling}, would also be crucial for scaling IRM to larger datasets and models.

\section*{Acknowledgements}

Y. C. would like to thank Chiheon Kim for helpful discussion and comments.


\bibliography{references}
\bibliographystyle{icml2020}

%%%%%%%%%%%%%%%%%%%%%%%%%%%%%%%%%%%%%%%%%%%%%%%%%%%%%%%%%%%%%%%%%%%%%%%%%%%%%%%
%%%%%%%%%%%%%%%%%%%%%%%%%%%%%%%%%%%%%%%%%%%%%%%%%%%%%%%%%%%%%%%%%%%%%%%%%%%%%%%
% APPENDIX
%%%%%%%%%%%%%%%%%%%%%%%%%%%%%%%%%%%%%%%%%%%%%%%%%%%%%%%%%%%%%%%%%%%%%%%%%%%%%%%
%%%%%%%%%%%%%%%%%%%%%%%%%%%%%%%%%%%%%%%%%%%%%%%%%%%%%%%%%%%%%%%%%%%%%%%%%%%%%%%
\clearpage
\appendix

\section{ColoredMNIST is a Special Case of Extended ColoredMNIST}
\label{app:original}

The original ColoredMNIST setup from \cite{arjovsky2019invariant} is a special case of Extended ColoredMNIST. 
It consists of $m=2$ environments and a test environment, all with label corruption probabilities $\eta_1 = \eta_2 = \eta_\mathrm{test} = 0.25$. 
The two training environments have coloring probabilities $p_1 = 0.2$ and $p_2 = 0.1$ respectively, while the test environment has $p_\mathrm{test} = 0.9$. 
Labels are collapsed into two classes, $y=0$ for digits 0-4 and $y=1$ for digits 5-9.
To color the image, the input image is represented in two channels, one for red and one for green. 
During training, the color of each input image is highly correlated to the binary label: 0 with green and 1 with red (with probability $1-p_e$ for each environment $e$).
At test time, this correlation is reversed: 0 with red and 1 with green (with probability $p_\mathrm{test}$).

\section{Extended ColoredMNIST Training Details}
\label{app:coloredmnist_hparams}

Our code\footnote{\url{https://github.com/kakaobrain/irm-empirical-study/tree/master/colored_mnist}} is based on the official implementation for \cite{arjovsky2019invariant}, and we chose our hyperparameters as follows.
The representation model is parametrized as a multi-layer perceptron (MLP) with ReLU activations, three layers, and an equal number ($h=256$) of hidden units in between layers, containing a total of 166,914 trainable parameters. 
The model is trained using gradient descent for 500 steps, with a fixed learning rate of $0.001$ and a L2 weight decay of $0.001$.
$\lambda$ is set to $10,000$ but is fixed to $1$ for the first $100$ steps.
Input MNIST images were subsampled from 28x28 to 14x14.
All reported accuracies are averaged over 10 random seeds, along with their standard deviations.

\section{Additional Extended ColoredMNIST Experiments}
\label{app:coloredmnist_additional}

In this section, we run additional experiments with different variations of Extended ColoredMNIST, complementing our results from \S\ref{sec:coloredmnist}.

\subsection{Is IRMv1 still effective when the invariant correlation is stronger than the spurious one?}
\label{sec:q2}

In the original ColoredMNIST experiment, by construction, the spurious correlation is stronger than the invariant correlation during training.
In reality, the spurious correlation might exist but its degree might not be as strong as the invariant correlation. 
We pose two questions here: (i) Does the weak spurious correlation still hurt ERM's generalization performance? (ii) If so, can IRMv1 help avoid this issue?

To answer these questions, we setup two Extended ColoredMNIST settings where the invariant correlation is stronger the spurious one. 
In the first setting, the spurious correlations for training environments are weakened to $p_1 = 0.45$ and $p_2 = 0.35$, with label corruption probabilities kept at $\eta_1 = \eta_2 = \eta_\mathrm{test} = 0.25$.
In the second setting, we remove label corruption, i.e. $\eta_1 = \eta_2 = \eta_\mathrm{test} = 0.0$, and the spurious correlations are unchanged from the original version, i.e. $p_1 = 0.2$ and $p_2 = 0.1$.
In both settings, the average spurious correlation ($1 - \frac{p_1 + p_2}{2}$) is now 0.15 lower than the invariant correlation ($1 - \eta$), and the gap between training environments ($p_1 - p_2$) is fixed to $0.1$.
Also, $p_\mathrm{test}$ is kept as $0.9$ in both settings.

\begingroup
\setlength{\tabcolsep}{2pt} % Default value: 6pt
\renewcommand{\arraystretch}{1.2} % Default value: 1
\begin{table}[t]
\centering
\small
\begin{tabular}{c | r r | r r}
\Xhline{1.1pt}
\multirow{2}{*}{\bf Algorithm} & \multicolumn{2}{c|}{\bf 25\% Label Corruption} & \multicolumn{2}{c}{\bf No Label Corruption} \\
                           & \multicolumn{1}{c}{\it Train} & \multicolumn{1}{c|}{\it Test} & \multicolumn{1}{c}{\it Train} & \multicolumn{1}{c}{\it Test} \\ \hline
ERM                        & $81.5 \pm 0.5$ & $61.6 \pm 2.0$ & $99.6 \pm 0.0$ & $\bf 92.7 \pm 0.2$ \\ 
IRMv1                      & $74.0 \pm 0.3$ & $\bf 71.6 \pm 0.9$ & \textsuperscript{\textdagger}$98.3 \pm 0.0$ & \textsuperscript{\textdagger}$91.0 \pm 0.3$ \\ \hline
Random                     & $50$ & $50$ & $50$ & $50$ \\
Optimal                    & $75$ & $75$ & $100$ & $100$ \\
Grayscale                  & $76.6 \pm 0.3$ & $71.6 \pm 0.5$ & $99.3 \pm 0.1$ & $97.9 \pm 0.1$ \\ 
\Xhline{1.1pt}
\end{tabular}
\caption{Accuracy (\%) on Extended ColoredMNIST, \emph{where the invariant correlation is stronger than the spurious one}. Averaged over 10 trials (mean $\pm$ standard deviation). \textsuperscript{\textdagger}Trained longer for 10,000 (x20) steps, around which both the ERM loss and the IRMv1 penalty stopped decreasing.}
\label{tbl:q2}
\end{table}
\endgroup

In Table \ref{tbl:q2}, we present train and test accuracies in the two settings. 
We also present random, oracle, and grayscale baselines in both settings.
With 25\% label corruption, we find that IRMv1 achieves the same accuracy (71.6\%) as the grayscale model, suggesting that \emph{it effectively ignores the spurious correlation and predicts as well as the same model trained without the spurious correlation.} 
In contrast, while ERM now achieves accuracy above chance (61.6\%), it fails to completely ignore the spurious correlation when making its prediction, as evidenced by its relative high train and low test accuracies.

\emph{With no label corruption, however, we find that ERM (92.7\%) outperforms IRMv1 (91.0\%)}, even though IRMv1 needed much more training steps (10,000) for the training loss and penalty to stop decreasing. 
One possible explanation is that, because the invariant correlation is ``too strong'', the difference in spurious correlations between training environments was comparatively not large enough for IRMv1 to exploit. 
This would suggest that IRMv1 is less effective when the invariant correlation is already strong enough, such as the case of no label corruption. 
Also, notice that ERM's performance is still lower than the grayscale model (97.9\%), suggesting that the spurious correlation is still problematic for ERM, although IRMv1 does not seem to solve this issue.

\subsection{Does IRMv1 work with more than two training environments?}
\label{sec:q4}

Some realistic datasets for IRM may contain examples sourced from many more environments than two.
When datasets are collected from multiple sources, which is common for many benchmark datasets, they are likely separable into many environments, each with a different degree of spurious correlation.

To examine the performance of IRMv1 with $m \geq 2$ environments, we follow the Extended ColoredMNIST data construction pipeline we described earlier to build datasets with $m = 3, 5, 10$ training environments. 
Each environment $e$ possesses a unique probability $p_e$ that the label is correlated to a specific color. 
In all cases, we set maximum and minimum values of $p_e$ to $0.3$ and $0.1$, respectively, and spread out the environments evenly.
For $m=3$ environments, we use $p_1 = 0.3$, $p_2 = 0.2$, and $p_3 = 0.1$. %$(p_1, p_2, p_3) = (0.3, 0.2, 0.1)$.
We also test out uneven gaps when $m=5$ using $p_1 = 0.3$, $p_2 = 0.25$, $p_3 = 0.17$, $p_4 = 0.15$, and $p_5 = 0.1$. % $(p_1, p_2, p_3, p_4, p_5) = (0.3, 0.25, 0.17, 0.15, 0.1)$
Note that the average of environment probabilities is always smaller than $\eta = 0.25$, meaning that ERM performs poorly ($<50$\%) in all of these environments.

\begingroup
\setlength{\tabcolsep}{4pt} % Default value: 6pt
\begin{table}[t]
\centering
%\small
\begin{tabular}{c | c | r r}
\Xhline{1.1pt}
\multirow{2}{*}{\bf Algorithm} & \multirow{2}{*}{\bf \# Envs.} & \multicolumn{2}{c}{\bf Accuracy} \\
                           &                              & \multicolumn{1}{c}{\it Train} & \multicolumn{1}{c}{\it Test}  \\ \hline
ERM                        & 2                            & $86.4 \pm 0.9$  & $28.5 \pm 3.8$  \\ 
IRMv1                      & 2                            & $72.0 \pm 0.5$  & $\bf 70.1 \pm 0.8$  \\ 
IRMv1                      & 3                            & $71.8 \pm 0.8$  & $69.6 \pm 1.4$  \\ 
IRMv1                      & 5                            & $72.2 \pm 1.2$  & $68.3 \pm 0.8$  \\ 
IRMv1                      & 5 (uneven)                   & $70.9 \pm 0.8$  & $68.5 \pm 1.5$  \\ 
IRMv1                      & 10                           & $72.2 \pm 0.9$  & $68.4 \pm 1.4$  \\ 
\Xhline{1.1pt}
\end{tabular}
\caption{Accuracy (\%) on Extended ColoredMNIST \emph{with multiple environments} ($m=2, 3, 5, 10$). In each setting, the maximum gap in coloring probabilities between training environments is 0.2 and their average is less than 0.25. Averaged over 10 trials (mean $\pm$ standard deviation).}
\label{tbl:q4}
\end{table}
\endgroup

Our results are summarized in Table \ref{tbl:q4}. 
Overall, we find that \emph{IRMv1 can achieve high test accuracy (68.3-70.1\%) with 3, 5, or 10 training environments, spread out evenly or unevenly}. 
Also, the performance seems slightly better for less numbers of environments, although not significantly.
We posit that the performance might degrade for more environments as the average gap between any two environments gets closer, because we fixed the maximum gap.

\subsection{Does IRMv1 work with multiple outcomes?}
\label{sec:q5}

Our final Extended ColoredMNIST experiment examines the performance of IRMv1 for multiple outcomes. 
This is important because many real-world datasets involve multi-dimensional outputs -- most notably, multi-class classification tasks require multi-dimensional logits as outputs. 
Yet, for the sake of clarity, the original formulation of IRMv1 in \cite{arjovsky2019invariant} focused on binary classification with sigmoidal logits, leading to a scalar output.
Here, we treat ColoredMNIST as a multi-class classification task, per the extended derivation of IRMv1 in \eqref{eqn:irmv1}.

Analogous to ColoredMNIST with two classes, we construct a $k$-class ColoredMNIST by assigning a unique color that is highly correlated to each output class during training and shifting it for the test set. 
To prevent introducing unwanted correlation structures, we assign a unique input channel for each color.
Note that this makes the first layer of the MLP contain more parameters for larger $k$.
We test four values of $k$: $k=2$ ($y=0$ for digits 0-4, $y=1$ for 5-9); $k=5$ ($y=0$ for 0-1, $y=1$ for 2-3, ..., $y=4$ for 8-9); and $k=10$ (each digit is its own class).

\begingroup
\setlength{\tabcolsep}{4pt} % Default value: 6pt
\begin{table}[t]
\centering
%\large
\begin{tabular}{c | c | r r}
\Xhline{1.1pt}
\multirow{2}{*}{\bf Algorithm} & \multirow{2}{*}{\bf \# Outcomes} & \multicolumn{2}{c}{\bf Accuracy} \\
                           &                              & \multicolumn{1}{c}{\it Train} & \multicolumn{1}{c}{\it Test}  \\ \hline
ERM                        & \multirow{4}{*}{2}           & $89.1 \pm 0.4$  & $19.6 \pm 1.0$  \\ 
IRMv1                      &                              & $71.4 \pm 0.8$  & $\bf 67.6 \pm 1.3$  \\ 
Random                     &                              & $50$            & $50$            \\ 
Grayscale                  &                              & $76.6 \pm 0.2$  & $71.6 \pm 0.4$  \\ \hline
ERM                        & \multirow{4}{*}{5}           & \textsuperscript{\ddag}$95.2 \pm 0.2$  & \textsuperscript{\ddag}$41.0 \pm 0.6$  \\ 
IRMv1                      &                              & $82.2 \pm 0.4$  & $\bf 62.0 \pm 2.4$  \\ 
Random                     &                              & $20$            & $20$            \\ 
Grayscale                  &                              & $73.2 \pm 0.2$  & $71.7 \pm 0.4$  \\ \hline
ERM                        & \multirow{4}{*}{10}          & \textsuperscript{\ddag}$92.6 \pm 0.2$  & \textsuperscript{\ddag}$39.2 \pm 0.9$  \\ 
IRMv1                      &                              & \textsuperscript{\textdagger}$83.4 \pm 0.5$  & \textsuperscript{\textdagger}$\bf 58.6 \pm 2.5$  \\ 
Random                     &                              & $10$            & $10$            \\ 
Grayscale                  &                              & $73.2 \pm 0.1$  & $71.9 \pm 0.5$  \\
\Xhline{1.1pt}
\end{tabular}
\caption{Accuracy (\%) on Extended ColoredMNIST \emph{with multiple outcomes} ($k=2, 5, 10$). Averaged over 10 trials (mean $\pm$ standard deviation).
\textsuperscript{\textdagger}Trained longer for 1,000 (x2) steps, around which both the ERM loss and the IRMv1 penalty stopped decreasing. \textsuperscript{\ddag}Trained longer for 5,000 (x10) steps, around which the ERM loss stopped decreasing.}
\label{tbl:q5}
\end{table}
\endgroup

Our results are summarized in Table \ref{tbl:q5}. 
Overall, we find that \emph{IRMv1 still generalizes significantly better than ERM on multiple outcomes ($k=5, 10$).}
This shows that the IRMv1 penalty for multi-dimensional outputs is still effective, so long as the underlying causal structure is preserved.
We do note that the test accuracy for IRMv1 degrades as more output classes are used, unlike the grayscale model that retains its test accuracy.
This suggests that, although effective, the IRMv1 penalty may become less effective when its squared gradients are summed over more dimensions.
One possibility is that the gradient norm penalty does not scale each of the $k$ dimensions adaptively, although it is unclear how to weigh each dimension properly.
Another possibility is that evaluating the gradient at the all-1 vector is problematic, although the IRMv1 derivation for least-squares in \cite{arjovsky2019invariant} suggests that this shouldn't be an issue.

\section{PunctuatedSST-2 Training Details}
\label{app:punctuatedsst2_hparams}

Our code\footnote{\url{https://github.com/kakaobrain/irm-empirical-study/tree/master/punctuated_sst2}} for PunctuatedSST-2 extends the data construction and training scripts for Extended ColoredMNIST.
For text pre-processing, we reference the SST-2 data pre-processing script from HuggingFace's \texttt{transformers}\footnote{\url{https://github.com/huggingface/transformers}} library.

For hyperparameters, we use the average of the 300-dimensional GloVe (\texttt{glove.6B.300d}) \cite{pennington2014glove} word vectors as inputs, and train a 3-layer (300-300-2) perceptron with ReLU activations. 
The model contains 181,202 trainable parameters, which is comparable to the MLP model used for ColoredMNIST.
We still train with full gradient descent for 500 steps, but use \texttt{hyperopt}\footnote{\url{https://github.com/hyperopt/hyperopt}} with tree-structured Parzen estimators (TPE) \cite{bergstra2015hyperopt} for a hyperparameter search over the learning rate (1e-3, 2e-3, 5e-3, 1e-2), L2 weight decay (1e-4, 5e-4, 1e-3, 1e-2), and the penalty weight $\lambda$ for IRMv1 (5k, 7.5k, 10k, 20k, 50k).
Across 50 hyperparameter configurations, we choose the one that gives the highest minimum accuracy over the three test sets (averaged over 5 trials).
We then use the best hyperparameter configuration to report the final mean accuracy.

%%%%%%%%%%%%%%%%%%%%%%%%%%%%%%%%%%%%%%%%%%%%%%%%%%%%%%%%%%%%%%%%%%%%%%%%%%%%%%%
%%%%%%%%%%%%%%%%%%%%%%%%%%%%%%%%%%%%%%%%%%%%%%%%%%%%%%%%%%%%%%%%%%%%%%%%%%%%%%%

\end{document}